\documentclass[10pt,twocolumn,letterpaper]{article}

\usepackage[english]{babel}
\usepackage[utf8x]{inputenc}
\usepackage[T1]{fontenc}

\usepackage[a4paper,top=2cm,bottom=2cm,left=2cm,right=2cm,marginparwidth=2cm]{geometry}

\usepackage[toc,page]{appendix}
\usepackage{tabularx}
\usepackage[dvipsnames]{xcolor}
\usepackage{amsmath}
\usepackage{amssymb}
\usepackage{graphicx}
\usepackage[colorinlistoftodos]{todonotes}
\usepackage[colorlinks=true, allcolors=blue]{hyperref}
\usepackage{natbib}
\title{
		\usefont{OT1}{bch}{b}{n}
		\normalfont \normalsize
        \normalfont \normalsize \textsc{CS 5100: Foundations of AI Course Project} \\ [10pt]
		\huge Improving Conditional VAE with Non-Volume Preserving transformations \\
}
\usepackage{authblk}
\author{Tuhin Subhra De}
\affil{Northeastern University}
\affil{\textit{de.t@northeastern.edu}}
\date{\vspace{-2ex}}
\begin{document}
\maketitle

\begin{abstract}
Variational Autoencoders and Generative Adversarial Networks remained the state-of-the-art (SOTA) generative models until 2022. Now they are superseded by diffusion based models. Efforts to improve traditional models have stagnated as a result. In old-school fashion, we explore image generation with conditional Variational Autoencoders (CVAE) to incorporate desired attributes within the images. VAEs are known to produce blurry images with less diversity; we refer to a method that solves this issue by leveraging the variance of the gaussian decoder as a learnable parameter during training. Previous works on CVAEs assumed that the conditional distribution of the latent space given the labels is equal to the prior distribution, which is not the case in reality. We show that estimating it using Non-Volume Preserving (NVP) transformations results in better image generation than existing methods by reducing the FID by 4\% and increasing log likelihood by 7.6\% compared to the previous cases \footnote{The code is publicly available at: \href{https://github.com/Gituhin/Conditional-VAE-NF}{https://github.com/Gituhin/Conditional-VAE-NF}}. 
\end{abstract}

\section{Introduction}
Connecting variational inference with deep learning led to the invention of VAE \cite{VAE}. It is a deep density model that can be used in probabilistic modeling and representation learning as they are both conceptually simple and are able to scale to very complex distributions and large datasets. Unlike general autoencoders, the core idea of a VAE is to project the input into the parameters of a latent space distribution. Then the latent space is sampled from these parameters and passed into the decoder for reconstruction of the original input. We see the Evidence Lower Bound (ELBO) of the observed data distribution in the next subsection.

\subsection{ELBO for VAE}
\label{subsec: ELBO derive-VAE}

Let \( x \) be the observed data and \( z \) the latent variable. The generative model is defined as:
\begin{equation}
    \label{eqn:joint-dist}
    p(x, z) = p(x|z)p(z) = p(z|x)p(x)
\end{equation}

We are interested in finding the marginal $p(z|x)$ or the likelihood of $z$ given $x$. From Eqn \ref{eqn:joint-dist}, we see that the marginal is given by:

\begin{equation}
    \label{eqn:marginal}
    p(z|x) = \frac{p(z, x)}{p(x)} = \frac{p(x, z)}{\int p(x, z) \, dz}
\end{equation}
Here, the denominator in Eqn \ref{eqn:marginal} is intractable. It means that there does not exist a closed form solution to the integral or it requires exponential time to compute by sampling methods \cite{Blei_2017} hence, making $p(z|x)$ as intractable itself. To solve this problem, we introduce a tractable family of distributions $Q$ to approximate the marginal. The optimal distribution from the family is $q^*(z|x) \in Q$ which minimizes the KL-Divergence between $q$ and $p$.
\begin{equation}
\label{eqn:VI-approx}
    q^*(z|x) = arg\min_{q\in Q} \mathrm{KL} \left( q(z|x) \, || \, p(z|x)\right)
\end{equation}

When this is expanded, yields the ELBO as the objective equation (See Appendix \ref{appndx:elbo-VAE} for derivation):
\begin{equation}
\label{eqn:ELBO-VAE}
    \log p(x) \geq \mathbb{E}_{q(z|x)} \left[ \log p(x|z) \right] - \mathrm{KL}(q(z|x) \| p(z))
\end{equation}

We interpret the individual terms of the objective as:

\begin{itemize}
    \item \textbf{Reconstruction Term:} \( \mathbb{E}_{q(z|x)}[\log p(x|z)] \) encourages the decoder to reconstruct \( x \) accurately from \( z \).
    \item \textbf{Regularization Term:} \( \mathrm{KL}(q(z|x) \| p(z)) \) penalizes divergence between the approximate posterior and the prior on latent space.
\end{itemize}

\textbf{Final VAE Objective}

The VAE is trained by maximizing the ELBO, or equivalently minimizing the negative ELBO. Maximizing the ELBO decreases our objective in Eqn \ref{eqn:VI-approx}:
\begin{equation}
    \label{eqn:loss-VAE}
    \mathcal{L}_{\text{VAE}} = -\mathbb{E}_{q(z|x)}[\log p(x|z)] + \mathrm{KL}(q(z|x) \| p(z))
\end{equation}

\subsection{ELBO for Conditional VAE}

\begin{figure}[!ht]
  \centering
  \includegraphics[width=0.45\textwidth]{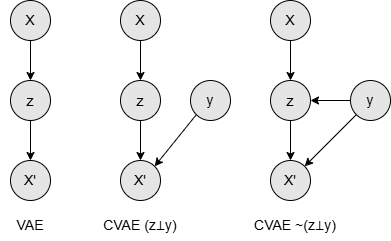}
  \caption{Graphical model for VAE and CVAE. The model at the center has its latent space independent of the labels. Whereas the model in the right has dependency. X is the input, y is the label/attribute and X' is reconstructed input.}
  \label{fig: VAE-CVAE model}
\end{figure}

Following the same process in above, we can derive the ELBO for the conditional VAE. Figure \ref{fig: VAE-CVAE model} illustrates the difference between normal VAE and conditional VAE. We see the only difference is the introduction of attribute variables $y$ with the input $x$. In this case the approximate posterior modifies the Eqn \ref{eqn:VI-approx} as:
\begin{equation}
    \label{eqn:CVI-approx}
    q^*(z|x, y) = arg\min_{q\in Q} \mathrm{KL} \left( q(z|x, y) \, || \, p(z|x, y)\right)
\end{equation}

We can expand the above equation following the rightmost model in the Bayesian network shown in Figure \ref{fig: VAE-CVAE model} and this gives us the ELBO as (See Appendix \ref{appndx:elbo-CVAE} for derivation):
\begin{equation}
    \label{eqn:obj-CVAE}
    \resizebox{0.9\hsize}{!}{$\log p(x, y) \geq \mathbb{E}_{q(z|x, y)} \left[ \log p(x|z, y) \right] - \mathrm{KL}(q(z|x, y) \| p(z|y))$}
\end{equation}

\textbf{Final Conditional VAE Objective}

The loss function is given by:
\begin{equation}
    \label{eqn:loss-CVAE}
    \resizebox{0.9\hsize}{!}{$\mathcal{L}_{\text{CVAE}} = -\mathbb{E}_{q(z|x, y)}[\log p(x|z, y)] + \mathrm{KL}(q(z|x, y) \| p(z|y))$}
\end{equation}

\subsection{Related Works}

The conditional $p(z|y)$ is difficult to estimate as again this does not have a closed form solution. In simple VAE, \cite{VAE} set the prior $p(z) = \mathcal{N}(0, I)$. This leads to a simple objective for the KL term in Eqn \ref{eqn:loss-VAE}. Previous works \cite{Attribute-CVAE-adobe}, \cite{CVAE-for-learned-img} ignore the marginal and set $p(z|y) = p(z)$ in the objective. We show that this does not inhibits modeling on the data but results in poor conditional reconstruction of the data. 

VAEs are known to produce blurry images with less diversity \cite{beta-VAE}. They use Lagrange multipliers to redefine the KL divergence term on the objective to be bounded by a small positive value. This gives rise to $\beta-$VAE. But this does not resolves the blurriness of image generation and lack of diversity.  \cite{sigma-VAE} also pointed out this issue. We use their method of leveraging the variance of the decoder as a parameter to incorporate diversity in generated images.

The rest of the article is organized as following Section 2 describes the methods to resolve the problems described above, Section 3 presents some visual and tabular results in a comparative way on performance of the proposed methodologies. Section 4 discusses the limitations and future work.
\section{Methodology}
We survey the methods to resolve the two major issues with CVAEs. The blurry image generation with lack of diversity and the estimation of the conditional $p(z|y)$ in Eqn \ref{eqn:loss-CVAE}.

\subsection{Optimal CVAE}
\cite{sigma-VAE} proposed a great yet simple method to remove the monotonicity in the generated images by the VAE. In general, the log likelihood or the reconstruction term in the objective is parametrized by a decoder which is Gaussian in nature. We can write the term with reconstructed input $\hat{x}$ and original input $x \in \mathbb{R}^P$ following $x|z, y \sim \mathcal{N}(\hat{x}, \, I)$ as:

\[
    -\ln p(x|z, y) = \log \mathcal{N}(x; \hat{x}, \, I)
\]
\[
\implies -\ln p(x|z, y) = \frac{1}{2} \| \hat{x} - x \|^2 + P \ln \sqrt{2\pi} 
\]
\[
\implies -\ln p(x|z, y) = \frac{P}{2} \mathrm{MSE} (x, \hat{x}) + c
\]
This results in generation of similar similar kind of images by the decoder due to the presence of unit variance. The authors replace the unit variance with a learnable parameter $\sigma$. Hence, the log likelihood now can be written as:

\begin{equation}
    \label{eqn:sigma-VAE-reconst-loss}
    -\ln p(x|z, y) = \log \mathcal{N}(x; \hat{x}, \, \sigma^2I)
\end{equation}
\[
\implies -\ln p(x|z, y) = \frac{1}{2\sigma^2} \| \hat{x} - x \|^2 + P \ln \sigma + c
\]
\[
\implies -\ln p(x|z, y) = \frac{P}{2\sigma^2} \mathrm{MSE} (x, \hat{x}) + P \ln \sigma+ c
\]

When allowing the variance to be learned by gradient descent during training the authors found that variance learned is sub-optimal. They propose an analytical solution to determine the optimal variance in Eqn \ref{eqn:sigma-VAE-reconst-loss}. The optimal variance will be the one that maximizes the log likelihood. Let the optimal variance be $\sigma^{*2}$. Then it can written that:

\begin{equation}
    \label{eqn:optimal-sigma}
    \sigma^{*2} = \mathrm{arg}\max_{\sigma}\mathcal{N}(x; \hat{x}, \, \sigma^2I)
\end{equation}
\[
    \implies\sigma^{*2} = \frac{1}{P} ||x-\hat{x}||^2 = \mathrm{MSE} (x, \hat{x})
\]
The result shown above is the intuition that the Maximum Likelihood Estimation of a variance of a distribution is the sample variance itself. This can also be derived by differentiation (Refer to Appendix \ref{appndx:optimal-sigma}). Substituting the optimal variance in Eqn \ref{eqn:sigma-VAE-reconst-loss}, we get the log likelihood or reconstruction loss as:
\begin{equation}
    \label{eqn:sigma-VAE-optimal-reconst-loss}
    \mathcal{L}_{\text{R}}=-\ln p(x|z, y) = \frac{P}{2} \ln \mathrm{MSE} (x, \hat{x})+ c
\end{equation} 
In the batch setting, the optimal variance would be simply the MSE loss, and can be updated after every gradient update for the other parameters of the decoder. In the mini-batch setting, we use a batchwise estimate of the variance computed for the current minibatch.

\subsection{Estimation of the conditional}
As we discussed previously, estimation of the conditonal $p(z|y)$ is not possible analytically. From the graphical model of CVAE shown in Figure \ref{fig: VAE-CVAE model}, we can write $p(z|y) = \int p(z|x, y)p(x)dx$. We need to marginalize over $x$ however, we have seen earlier that calculating $p(x)$ is intractable. 

\subsubsection{Single stage normalizing flows}
\label{subsection:NF}
In order to tackle this, \cite{CVAE-TTS} introduced the concept of normalizing flows for estimating this. A normalizing flow describes the transformation of a probability density through a sequence of invertible mappings. By repeatedly applying the rule for change of variables, the initial density ‘flows’ through the sequence of invertible mappings. At the end of this sequence we obtain a valid probability distribution and hence this type of flow is referred to as a normalizing flow (NF) \cite{VI-with-NF}. Let \( A \) and \( B \) be random variables related by a mapping  
\( f : A \rightarrow B \) such that  

\[
B = f(A) \quad \text{and} \quad A = f^{-1}(B)
\]
Then the probability density function of \( A \) for $a\in A$ and $b \in B$ is given by:
\[
p_A(a) = p_B(b) \left| \det \left( \frac{\partial b}{\partial a} \right) \right|
\]
\[
\implies p_A(a) = p_B(f(a)) \left| \det \left( \frac{\partial f(a)}{\partial a} \right) \right|
\]

The authors in \cite{CVAE-TTS} parameterize the marginal by a transformed normal distribution whose parameters are obtained from the labels. Let $z|y \sim \mathcal{N}(\mu_p, \, \sigma_p)$ with a transformation function $g = f(z)$. Then the marginal will be given by:

\begin{equation}
    \label{eqn:marginal-NF}
    p(z|y) = \mathcal{N}(f(z); \mu_p, \, \sigma_p) \left| \det \left( \frac{\partial f(z)}{\partial z} \right) \right|
\end{equation}

Where $\mu_p = \mu_p(y), \, \sigma_p = \sigma_p(y)$ are learnable functions that take the labels as inputs and output the parameters of the distribution. Calculating the determinant of the Jacobian above is computationally expensive given the complexity of the transformation function $f$ and the dimensionality of $z \in \mathbb{R}^D$. Hence, the authors choose some specific functions whose determinant is of unit value (say identity function) or the transformation has the volume preserved. This relieves the KL divergence term from the determinant term.

\subsubsection{Real Non-Volume Preserving transformation}
However, the challenge of Jacobian calculation was easily solved by \cite{NVP}. They estimated probability density by non-volume preserving (NVP) transformations or the determinant is not of unit value. They redesigned the transformation function using an affine coupling layer. Let $s$ and $t$ be some transformation on $z$. Let $g = f(s(z), t(z))$, then for a $d < D$, the output $g$ of an affine coupling layer follows the equations:
\begin{align*}
    g_{1:d} &= z_{1:d} \\
    g_{d+1:D} &= z_{d+1:D} \odot \exp\left(s(z_{1:d})\right) + t(z_{1:d})
\end{align*}

Here:
- \( \odot \) denotes element-wise multiplication.
- \( s(\cdot) \) and \( t(\cdot) \) are learnable functions (e.g., neural networks).
The Jacobian of the above transformation can be easily calculated since it forms an upper triangular matrix as shown:

\[
\frac{\partial g}{\partial z} =
\begin{bmatrix}
I_d & 0 \\
\frac{\partial g_{d+1:D}}{\partial z_{1:d}} & \operatorname{diag}\left(\exp\left(s(z_{1:d})\right)\right)
\end{bmatrix}
\]

where \( \operatorname{diag}(\exp(s(z_{1:d}))) \) is the diagonal matrix whose diagonal elements correspond to the vector \( \exp(s(z_{1:d})) \). Given that this Jacobian is triangular, we can efficiently compute its determinant as:

\begin{equation}
    \label{eqn:det-NVP}
    \left| \frac{\partial g}{\partial z} \right| = \exp\left( \sum_j s(z_{1:d})_j \right)
\end{equation}
Since computing the Jacobian determinant of the coupling layer operation does not involve computing the Jacobian of \( s \) or \( t \), those functions can be arbitrarily complex. We can implement them as Multi-layered neural networks.

\subsection{Simplified Objective}
The KL Divergence term $\mathrm{KL}(q(z|x, y) \| p(z|y))$ can be calculated separately (See Appendix \ref{appndx:KL-from-NF} for more details). It is given by:

\[
   \mathcal{L}_{\text{KL}} = \log \sigma_p-\log \sigma_q -\frac{(f(z)-\mu_p)^2}{2\sigma_p^2} - \log \left| \det \left( \frac{\partial f(z)}{\partial z} \right) \right|
\]
Where $\mu_q, \, \sigma_q$ are the mean and variance of Encoder $q$. Combining the methods from the above two subsections we can write the final objective as:
\begin{equation}
    \label{eqn:loss-CVAE-final}
    \mathcal{L}_{\text{CVAE}} = \mathcal{L}_{\text{R}} + \mathcal{L}_{\text{KL}}
\end{equation}
\section{Experiments and Results}

\subsection{Experiments}

\begin{figure}[!ht]
  \centering
  \includegraphics[width=0.45\textwidth]{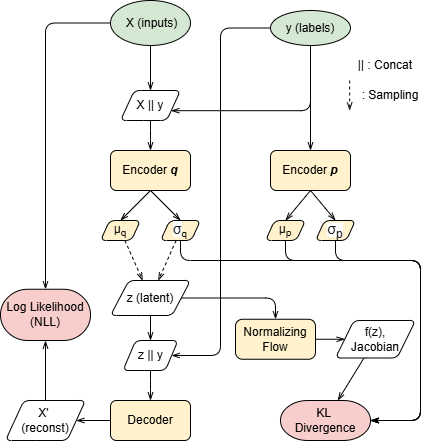}
  \caption{Flow of the $\sigma-$ CVAE (NVP) model during training. The blocks in yellow are trainable. Blocks in red depict the loss functions.}
  \label{fig: model-flow-training}
\end{figure}

We experiment in three settings:
\begin{itemize}
    \item \textbf{Gaussian CVAE:} CVAE with Gaussian decoder where the variance of the decoder is kept as 1 and not learned. Typically, the vanilla CVAE and latent space is dependent on the labels.
    
    \item \textbf{$\sigma-$CVAE (non-NVP)}: The scenario where the variance of the decoder is optimal as described and the latent space $z$ is independent of the labels. This sets $p(z|y) = p(z)$ the in objective equation.
    
    \item \textbf{$\sigma-$CVAE (NVP)}: The variance of the decoder is optimal and the latent space $z$ is dependent on the labels. $p(z|y)$ is estimated using NVP as described above.
\end{itemize}

The architectures of the models used are as follows:
\begin{itemize}
    \item \textbf{Encoder \textit{q}}: We use a convolutional neural network based encoder. This consists of four 2D-convolutional layers followed by one affine transformation for outputting the distribution parameters for the approximate posterior $q$.
    \item \textbf{Encoder \textit{p}}: The encoder for conditional $p$ comprises of two 1D-convolutional layers followed by one affine transformation.
    \item \textbf{Decoder}: It consists the same architecture as the Encoder \textit{q}, but in a way that upscale images i.e. four 2D-Convolutional Transpose layers.
    \item \textbf{NVP}: The functions $s$ and $t$ as described in subsection \ref{subsection:NF} are MLP with 2 hidden layers.
\end{itemize}

For incorporating the labels with the inputs and with the latent space, we simply concatenate them and pass into the models. Fig \ref{fig: model-flow-training} shows a representation of the data flow through the models.

For our experiments, we use the Celeb-A dataset \cite{CelebA}. This contains about 200k facial RGB images with 40 binary facial attributes describing the presence of certain facial features (e.g. blonde hair, makeup etc). We consider the images as our inputs and the binary attributes as the labels.

The entire models are trained on 160k images, each having spatial dimensions reduced to 86 $\times$ 86 with some random augmentations viz. randomly flipping horizontally, rotating by 10 degrees etc. With Nvidia's L4 GPU, the training time varies based on the early stopping method. Early stopping is based on the evaluation metrics (NLL) on the test set, which consists of about 40k images.

\subsection{Results}

\begin{table}[!ht]
    \centering
    \begin{tabularx}{0.45\textwidth}{c|c|c|c} 
        \hline
        Model & NLL & \begin{tabular}{@{}c@{}}FID \\ (Recon)\end{tabular} & \begin{tabular}{@{}c@{}}FID \\ (Sampled)$\downarrow$\end{tabular}\\ \hline
        Gaussian VAE & -32.95  &  389.20& 389.06\\
        $\sigma-$ VAE & -48.61 & 107.83 & 166.07\\
        $\sigma-$ VAE (NVP) & \textbf{-52.32}  & \textbf{107.24} & \textbf{159.13}\\ \hline
    \end{tabularx}
    \caption{Metrics of VAEs under given scenarios. NLL stands for Negative Log Likelihood on the reconstructed data (lower is better). FID (Recon) is the score on reconstructed images from the test set, whereas FID (Sampled) is the score on randomly sampled images and the test set. (See text for details)}
    \label{tab:NLL-FID metrics}
\end{table}
We compare the negative log likelihood (NLL) and the Fréchet Inception Distance (FID) \cite{FID}, \cite{FID-github-repo} score for all three settings. Table \ref{tab:NLL-FID metrics} shows the comparative metrics for all three scenarios. The NLL is calculated on the test set during the training. There are two FID scores, both of which are calculated post training. FID (Recon) or reconstructed is calculated between the reconstructed images test set images and the original test set. FID (Sampled) is calculated between the sampled images from random latent spaces with the attributes from the test set and the original test set images. We observe that CVAEs with approximated conditionals outperform all the other CVAEs in the settings.

\begin{figure}[!ht]
  \centering
  \includegraphics[width=0.48\textwidth]{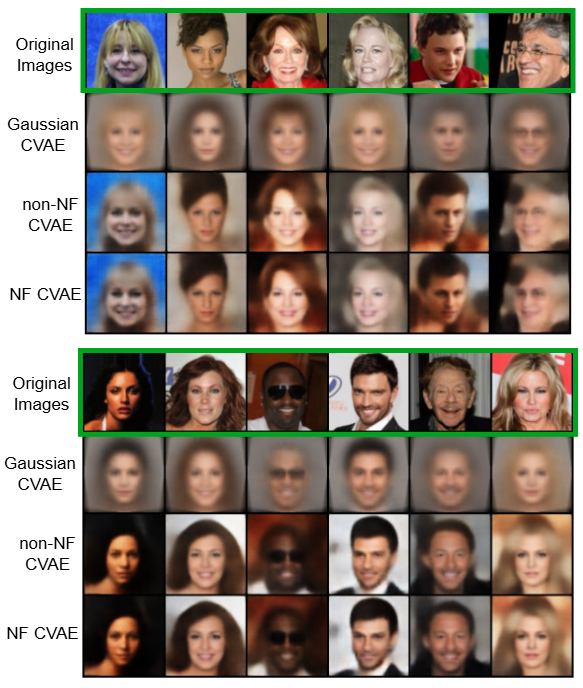}
  \caption{Reconstructions of the images from the test set by the models after training. The top row of each section marked in green border are the original images.}
  \label{fig: testset-recon}
\end{figure}

Some of the reconstructions from the test set are shown in Fig \ref{fig: testset-recon}. The problem of blurry image generation in the case of Gaussian CVAEs is pertinent in the results. Whereas optimal sigma CVAEs capture the variance in the images very well. The quality of reconstructions from non-NVP and NVP CVAEs are almost indistinguishable, however later in inference stage we will observe that NVP-CVAEs outrun non-NVP model visually.

\subsection{Inference}
\begin{figure}[!ht]
  \centering
  \includegraphics[width=0.25\textwidth]{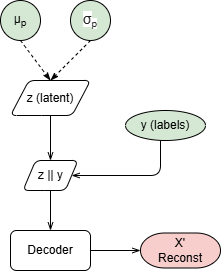}
  \caption{Flow of the $\sigma-$ CVAE (NVP) model during inference or sampling a random image with some labels.}
  \label{fig: model-flow-inference}
\end{figure}

Post training, we sample randomly generated images from the models. For the NVP scenario, since during the training the generated latent space already has a prior information of the labels, hence while inferencing we incorporate the information of labels by the transformation on latent as $\hat{z} = \mu_p + z*\sigma_p$ where $z \sim \mathcal{N}(0, I)$. Fig \ref{fig: model-flow-inference} describes the flow of inference through the models. The distribution parameters of encoder $p$ already have some information about the labels hence, they can be used to transform the randomly sampled latent space. For the rest of the scenarios, the latent space is directly sampled from standard Gaussian noise and passed into the decoder.

\begin{figure}[!ht]
  \centering
  \includegraphics[width=0.48\textwidth]{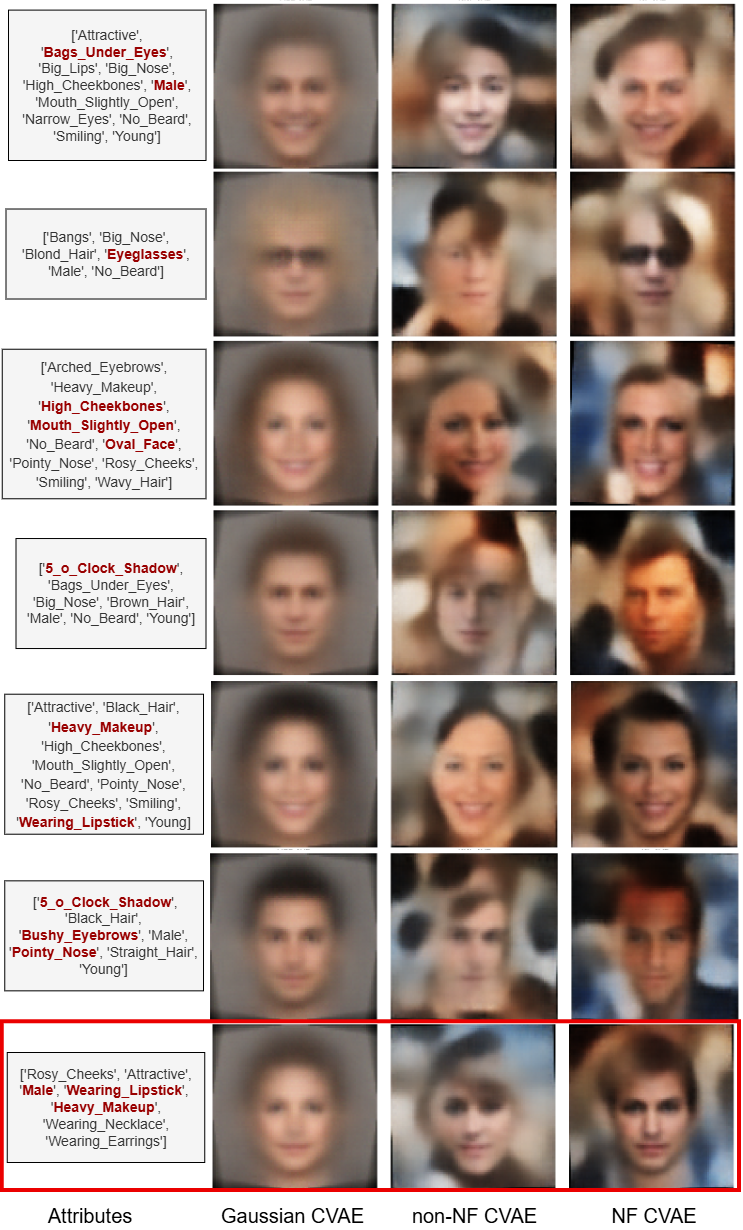}
  \caption{Comparison between the random images generated by the models under the described scenarios. The left text box contains the attributes on which the images were conditioned on. The ones in bolder \textcolor{Mahogany}{dark maroon} font are displayed prominently by NVP-CVAE. The last row in \textcolor{red}{red} border shows generation from attributes that might not be present in real life or during training.}
  \label{fig: generated-images-grid}
\end{figure}

Next, we obtain a few sampled images from all the three scenarios. Fig \ref{fig: generated-images-grid} presents some of the images generated by the models. The general observation that we get from these is that the NVP-CVAE excels in capturing the attributes as opposed to other models. They even recreate the learned representations of the attributes. The last row in the figure (highlighted in \textcolor{red}{red} rectangle) shows the efficient implementation of the attributes into the data which might not have been part of the training data or real life. A male wearing lipstick and heavy makeup is unlikely to be in the entire dataset.

\section{Limitations and Discussions}
To summarize, the objective of this work is not to generate high quality images like the SOTA models including stable diffusion \cite{SDXL}, vision-language models like Imagen-3 \cite{imagen3} etc, but in an old-school manner to focus on the process of estimating the model parameters using different statistical techniques. There are several limitations to this work. In order to align the attributes to the patches of images a cross-attention mechanism can be introduced between them. Even there can be dependencies within the attributes itself e.g. the one attribute with `heavy makeup' might be positively correlated with `lipstick' or `young' and negatively with `male'. This gives rise to possibility of self attention acting as a useful mechanism as well \cite{t2img_attention}.

Observing the background in the inference during randomly sampled images, it is quite evident that the latent space also includes some of the background information in it. This signifies that detecting the face in the foreground in a dedicated way through segmentation can help in better control over image generation with custom backgrounds.

Transposing convolutions as a up-scaling method used to go from the latent space to higher dimension might not be the best way. This obviously leads to blurry representations in Gaussian VAEs as seen previously. Adding some extra penalization in the intermediate reconstructions might help in sharper foreground object and background boundaries.
\section*{Conclusion}
In this work, we survey the methods to resolve the two major issues seen the VAEs. One being the blurry image generation with little to no variance. The other one being calculation of the conditional distribution of the latent space given the labels. We observe that finding the optimal variance of the decoder analytically gives the best quality of image generation. Estimation of the conditional distribution using normalizing flows assists the model to better incorporate the information of attributes during inference. This also leads to better FID score on the test set.

\bibliography{references}
\appendix
\section{Derivation of the Evidence Lower Bound (ELBO)}
\subsection{ELBO for VAE}
\label{appndx:elbo-VAE}
The optimal distribution from the family of distribution satisfying the equation:
\[
    q^*(z|x) = arg\min_{q\in Q} \mathrm{KL} \left( q(z|x) \, || \, p(z|x)\right)
\]
When we expand the RHS of the expression above, we get:
\[
\mathrm{KL}(q(z|x) \| p(z|x)) = \mathbb{E}_{q(z|x)} \left[ \log \frac{q(z|x)}{p(z|x)} \right]
\]
\[
\implies \mathbb{E}_{q(z|x)} \left[ \log \frac{q(z|x)}{p(z|x)} \right] = \mathbb{E}_{q(z|x)} \left[ \log \frac{q(z|x)p(x)}{p(x|z)p(z)} \right]
\]
\[
= \log p(x) + \mathbb{E}_{q(z|x)} \left[ \log q(z|x) - \log p(x|z) - \log p(z) \right]
\]

Rearrange to isolate \( \log p(x) \) as it is independent of the expectation under $q$:

\begin{equation*}
\begin{split}
        \log p(x) = \mathbb{E}_{q(z|x)} \left[ \log p(x|z) \right] - \mathrm{KL}(q(z|x) \| p(z)) + \\
        \mathrm{KL}(q(z|x) \| p(z|x))
\end{split}
\end{equation*}

Dropping the final KL term (as KL divergence > 0) yields the Evidence Lower Bound (ELBO) as the objective equation \ref{eqn:ELBO-VAE}:
\begin{equation*}
    \log p(x) \geq \mathbb{E}_{q(z|x)} \left[ \log p(x|z) \right] - \mathrm{KL}(q(z|x) \| p(z))
\end{equation*}

The ELBO can also be reached using Jensen's inequality upon maximizing the log likelihood on the joint data and latent distribution $p(x, z)$.

\subsection{ELBO for Conditional VAE}
\label{appndx:elbo-CVAE}
The optimal distribution from the family of distribution satisfying the equation:
\[
    q^*(z|x, y) = arg\min_{q\in Q} \mathrm{KL} \left( q(z|x, y) \, || \, p(z|x, y)\right)
\]
From the rightmost graphical model in the Figure \ref{fig: VAE-CVAE model}, we can write the joint distribution as:
\[
    p(x, y, z) = p(z|x, y)\,p(x)\,p(y)
\]
Using the Bayesian rule we can write the conditional marginal as:
\[
    p(z|x, y) = \frac{p(x|z, y)\,p(z|y)\,p(y)}{p(x, y)}
\]
Where the denominator $p(x, y) = \int p(x, y, z) dz$ is again intractable. For simplicity we refer $\mathbb{E}_{q(z|x, y)}$ as $\mathbb{E}_q$. Using this identity we can expand the objective as:
\[
\mathrm{KL}(q(z|x, y) \| p(z|x, y)) = \mathbb{E}_{q(z|x, y)} \left[ \log \frac{q(z|x, y)}{p(z|x, y)} \right]
\]
\[
\implies \mathbb{E}_q \left[ \log \frac{q(z|x, y)}{p(z|x, y)} \right] = \mathbb{E}_q \left[ \log \frac{q(z|x, y)p(x, y)}{p(x|z, y)p(z|y)p(y)} \right]
\]
\begin{equation*}
    \begin{split}
        \implies \mathbb{E}_q \left[ \log \frac{q(z|x, y)}{p(z|x, y)} \right] = \mathbb{E}_q \left[\log\frac{q(z|x, y)}{p(z|y)}\right] \\ -\mathbb{E}_q \log p(x|y, z) - \log p(x, y) - \log p(y)
    \end{split}
\end{equation*}

\begin{equation*}
    \begin{split}
        \implies \mathbb{E}_q \left[ \log \frac{q(z|x, y)}{p(z|x, y)} \right] = \mathrm{KL} \left( q(z|x, y) \, || \, p(z| y)\right) \\ -\mathbb{E}_q \log p(x|y, z) - \log p(x, y) - c
    \end{split}
\end{equation*}

Now we have:
\begin{equation*}
    \begin{split}
        \mathrm{KL}(q(z|x, y) \| p(z|x, y)) = \mathrm{KL} \left( q(z|x, y) \, || \, p(z| y)\right) \\ -\mathbb{E}_q \log p(x|y, z) - \log p(x, y) - c
    \end{split}
\end{equation*}
And as $\mathrm{KL}(q(z|x, y) \| p(z|x, y))>0$, we can rewrite the above equation to take the form of Eqn \ref{eqn:obj-CVAE} or ELBO for CVAE:
\begin{equation*}
    \begin{split}
         \log p(x, y) \geq \mathbb{E}_q \log p(x|z, y) - \mathrm{KL} \left( q(z|x, y) \, || \, p(z| y)\right)
    \end{split}
\end{equation*}

For the middle graphical model in the middle of Fig \ref{fig: VAE-CVAE model} which has the latent independent of the labels. The objective equation becomes:
\begin{equation*}
    \log p(x, y) \geq \mathbb{E}_{q(z|x)} \log p(x|z, y) - \mathrm{KL} \left( q(z|x) \, || \, p(z)\right)
\end{equation*}

\section{Optimal Variance of Decoder}
\label{appndx:optimal-sigma}
With the given equation:
\[
    \sigma^{*2} = \mathrm{arg}\max_{\sigma}\mathcal{N}(x; \hat{x}, \, \sigma^2I)
\]
\[
    \implies -\log \mathcal{N}(x; \hat{x}, \, \sigma^2I) = P\log \sigma + \frac{\left\lVert\hat{x}-x\right\rVert^2}{2\sigma^2} +c
\]
Where $P$ is the dimensionality of $x$. To maximize the objective, we take the first order derivative w.r.t $\sigma$ and equate it to 0.
\[
    \implies\frac{\partial \mathcal{\log N}}{\partial\sigma} = \frac{P}{\sigma} - \frac{\left\lVert\hat{x}-x\right\rVert^2}{\sigma^3} = 0
\]
\[
    \implies \sigma^2 =  \frac{\left\lVert\hat{x}-x\right\rVert^2}{P} \;\; \text{; since $\sigma \neq 0$}
\]


\section{Calculation of KL from Normalizing Flow}
\label{appndx:KL-from-NF}
The KL-Divergence is given by  $\mathrm{KL}(q(z|x, y) \| p(z|y))$, where
\begin{equation*}
    p(z|y) = \mathcal{N}(f(z); \mu_p, \, \sigma_p) \left| \det \left( \frac{\partial f(z)}{\partial z} \right) \right|
\end{equation*}
\[
\implies \mathrm{KL}(q(z|x, y) \| p(z|y)) = \mathbb{E}_{z\sim q(z|x, y)} \left[ \log \frac{q(z|x, y)}{p(z|y)} \right]
\]

\[
    \resizebox{0.95\hsize}{!}{$= \mathbb{E}_q \log q(z|x, y) - \mathbb{E}_q \left [-\log \sigma_p-\frac{(f(z)-\mu_p)^2}{2\sigma_p^2} + \log \det \left|\left( \frac{\partial f(z)}{\partial z}\right)\right|\right]$}
\]
\[
\resizebox{0.95\hsize}{!}{$=-\log \sigma_q + \log \sigma_p - \mathbb{E}_q \left [\frac{(f(z)-\mu_p)^2}{2\sigma_p^2} + \log \det \left|\left( \frac{\partial f(z)}{\partial z}\right)\right|\right]+c$}
\]

Expectation of log of a Gaussian is $\log \sigma +c$, hence $\mathbb{E}_q \log q(z|x, y) = -\log\sigma_q+c$ . A closed form solution does not exists for expectation above, hence we sample $z\sim q(z|x, y)$ and average out the term. Doing this makes the final KL-Divergence loss as:
\[
   \mathcal{L}_{\text{KL}} = \log \sigma_p-\log \sigma_q -\frac{(f(z)-\mu_p)^2}{2\sigma_p^2} - \log \left| \det \left( \frac{\partial f(z)}{\partial z} \right) \right|
\]

\end{document}